\documentclass[sigconf, 9pt]{acmart}

\AtBeginDocument{
  \providecommand\BibTeX{{
    Bib\TeX}}}

\copyrightyear{2026}
\acmYear{2026}
\setcopyright{cc}
\setcctype{by}
\acmConference[ICCAD '26]{IEEE/ACM International Conference on Computer-Aided Design}{November 08--12, 2026}{San Jose, CA, USA}
\acmBooktitle{IEEE/ACM International Conference on Computer-Aided Design (ICCAD '26), November 08--12, 2026, San Jose, CA, USA}
\acmDOI{10.1145/3831252.3834100}
\acmISBN{979-8-4007-2873-0/2026/11}

\usepackage{subcaption}
\usepackage{pifont}
\usepackage{multirow}
\usepackage{algorithm}
\usepackage{algpseudocode}
\usepackage[table]{xcolor}
\usepackage{bm}
\usepackage{enumitem}

\AtBeginEnvironment{algorithmic}{\small}

\renewcommand{\footnoterule}{\hrule width 0.25\textwidth height 0.5pt  \kern 3.5pt}

\definecolor{mycolor}{HTML}{E5E5FF}

\def\BibTeX{{\rm B\kern-.05em{\sc i\kern-.025em b}\kern-.08em
    T\kern-.1667em\lower.7ex\hbox{E}\kern-.125emX}}

\begin{document}

\title[MicroEvo: Knowledge-Guided LLM Sampling for Efficient Microarchitecture Design Space Exploration]{%
MicroEvo: Knowledge-Guided LLM Sampling for Efficient Microarchitecture Design Space Exploration
}

\author{\large
    Jia Xiong\textsuperscript{\rm \dag 1,2}, 
    Runkai Li\textsuperscript{\rm \dag 1,2},
    Chenxu Niu\textsuperscript{\rm 3},
    Guangyuan Gao\textsuperscript{\rm 1,2},
    Changwen Xing\textsuperscript{\rm 1,2},
    Yifan Zhang\textsuperscript{\rm 1,2}, 
    Xinlai Wan\textsuperscript{\rm 1,2},\\
    Jieran Cui\textsuperscript{\rm 1,2},
    Chen Bai\textsuperscript{\rm 4},
    Yusheng Hua\textsuperscript{\rm 5},
    Ying Wang\textsuperscript{\rm 6},
    Ming Ling\textsuperscript{\rm 1},
    Xi Wang\textsuperscript{\rm $\ast$1,2},
    Tao Xie\textsuperscript{\rm 1,4,7} \\
     \textsuperscript{\rm 1} Southeast University  \textsuperscript{\rm 2} National Center of Technology Innovation for EDA \\
     \textsuperscript{\rm 3} NVIDIA Corporation
     \textsuperscript{\rm 4} Fudan University
     \textsuperscript{\rm 5} Nanjing University of Posts and Telecommunications \\
     \textsuperscript{\rm 6} Institute of Computing Technology, Chinese Academy of Sciences 
     \textsuperscript{\rm 7} Peking University \& Beijing Tongming Lake Center  \\
     [-0.6ex]
}

\renewcommand{\shortauthors}{Jia Xiong et al.}

\begin{abstract}

Microarchitecture design space exploration suffers from expansive search spaces and expensive PPA evaluation, leaving only a small simulation budget for design decision-making. Existing methods perform blind search without considering microarchitectural dependencies and fail to learn from the iterative search effectively, leading to wasted evaluations and weak Pareto convergence. In this paper, we propose \textbf{MicroEvo}, a knowledge-guided framework that couples off-the-shelf LLMs with Monte Carlo Tree Search (MCTS) for multi-objective microarchitecture optimization. MicroEvo combines LLM-driven evolutionary operators, a Pareto-aware tree policy that balances Pareto contribution and diversity, an active knowledge accumulation mechanism that extracts and reuses optimization insights, and state-aware directives that adapt the search behavior online. Experiments show that MicroEvo improves Pareto-front quality by up to \textbf{36.2\%} over NSGA-II and achieves \textbf{10.6$\bm\times$} higher search efficiency, and also demonstrates strong scalability to a complex industrial-scale core. The code repository is available at: \url{https://github.com/GEAR-SEU/MicroEvo-ICCAD-26}.

\end{abstract}


\maketitle

\begingroup
\renewcommand{\thefootnote}{\dag}
\footnotetext{Equal Contribution.}
\endgroup

\begingroup
\renewcommand{\thefootnote}{$\ast$}
\footnotetext{Corresponding Author: xi.wang@seu.edu.cn}
\endgroup

\section{Introduction}
\label{sec:Introduction}
Emerging applications are increasingly driving hardware design toward workload-specific microarchitectures \cite{cpu_autodriving, cpu_mobile, samsung}. Under this trend, designers need to search over a broad space of feasible microarchitecture configurations to identify desirable performance, power, and area (PPA) trade-offs. However, the design space of modern computer architecture is typically high-dimensional and discrete, making design space exploration (DSE) particularly challenging. The impact of microarchitecture parameters on PPA is highly nonlinear, and even small parameter changes can alter the underlying dataflow logic \cite{ArchExplorer}. Moreover, PPA evaluation often has a long turnaround time, preventing timely feedback to adjust optimization decisions \cite{APPLE-DSE}. As a result, under the tight schedules of practical processor development, scarce evaluation opportunities must be used as efficiently as possible to approach the Pareto front.

To address this expensive multi-objective optimization (MOO) problem, industrial practice often relies on the prior knowledge of architects to directly propose candidate designs \cite{amd_processor}, but such approaches are inevitably constrained by individual bias and limited expertise. Earlier methods attempt to accelerate PPA estimation using analytical models \cite{analysis_isca, analysis_micro}. However, they require substantial modeling effort and suffer from poor transferability as microarchitectures evolve. Recent studies propose machine learning-based surrogate models to predict tool evaluations, reducing both modeling difficulty and evaluation cost \cite{boomexplorer, modse, MetaDSE, IT-DSE, trendse}. Although these methods have significantly improved search efficiency, the limited accuracy of absolute PPA prediction in the small-sample regime of microarchitecture DSE still introduces deviations from ground truth, leading to suboptimal results \cite{aldse}. Consequently, existing methods remain insufficient for the sampling-efficiency and Pareto-coverage requirements of modern microarchitecture DSE.

These limitations suggest that, in expensive microarchitecture DSE, the key challenge is \textit{not only better PPA prediction, but also better sampling}. When only a small number of design evaluations can be afforded, effective DSE must \textbf{make each sampling count} by proposing informative and high-quality candidates. Lacking human prior knowledge, conventional MOO strategies (e.g., evolutionary algorithms and Bayesian optimization) rely on probabilistic sampling for blind search. Consequently, they often waste the simulation budget on inefficient configurations \cite{EC_Review}. Large language models (LLMs) are pretrained on massive corpora containing textbooks, design documents, and other forms of microarchitecture knowledge \cite{LEMOE}. This factor equips them with the potential to generate designs that inherently respect architectural constraints and dependencies among microarchitecture components, thereby improving sample quality. To apply an LLM in microarchitecture DSE, existing methods iteratively feed PPA feedback to the LLM to guide exploration \cite{LEMOE, ChatA2, ChatArch, ChatDSE}. However, these studies merely learn superficial numerical information from the PPA data. They fail to fully exploit the knowledge-driven generation and reasoning capabilities of the LLM to provide informative guidance for each iteration \cite{eoh, FunSearch, MCTS-AHD, reevo}. Consequently, the LLM acts more as a black-box numerical optimizer, thereby undermining the reliability of search decisions.

Existing data-driven DSE methods often exhibit poor sample efficiency, leading to weak optimization performance when few evaluations are affordable. 
To address this issue, we propose \underline{Evo}lution of \underline{Micro}architecture (\textbf{MicroEvo}), a knowledge-guided framework for microarchitecture DSE.
The key idea is to combine the reasoning ability of LLMs with the structured search behavior of Monte Carlo Tree Search (MCTS), enabling MicroEvo to continuously learn optimization experience from DSE and evolve its search strategy. Instead of treating the LLM as a one-shot sampler or a black-box predictor, MicroEvo places it in an evolutionary search loop and equips it with a Pareto-aware tree policy, optimization knowledge, and state-aware search directives. This design transforms blind probabilistic search into informed sampling that respects microarchitectural dependencies. Accordingly, MicroEvo can better balance exploration and exploitation and more effectively approach the Pareto front in the expensive microarchitecture DSE.

Our main contributions are summarized as follows:
\vspace{-2pt}
\begin{itemize}[leftmargin=*]
    \item We propose an \textbf{LLM-driven evolutionary microarchitecture DSE framework} that integrates MCTS with \textit{knowledge-guided operators}. This framework achieves informed local refinement and global design refactoring for sample-efficient exploration.

    \item We propose \textbf{Pareto-UCT}, an evolutionary decision tree policy for multi-objective microarchitecture optimization. It selects expansion nodes according to both Pareto quality and coverage, making MCTS more suitable for global PPA-oriented DSE.

    \item We introduce \textbf{Active Knowledge Accumulation (AKA)}, which extracts reusable optimization knowledge from explored designs through \textit{Pareto analysis} and parent-child \textit{pairwise analysis}. AKA maintains a utility-aware memory to manage and retrieve useful insights for subsequent expansions.

    \item We design \textbf{State-Aware Directive (SAD)} that monitors real-time DSE progress and dynamically switches among \textit{exploit}, \textit{balanced}, and \textit{explore} modes. This design allows the LLM operators to adapt their modification behavior to the current search state, helping the search alleviate stagnation and local optima.

    \item Extensive experiments show that MicroEvo consistently achieves better Pareto quality under the same evaluation, and remains effective even in low-budget settings, with up to \textbf{36.2\%} improvement in hypervolume and \textbf{10.6$\bm\times$} search efficiency over NSGA-II.
\end{itemize}

\vspace{-5pt}

\section{Background \& Motivation}
\label{sec:Background}

\begin{figure}[t]
\centering
\subfloat[]{\includegraphics[width=4.15cm]{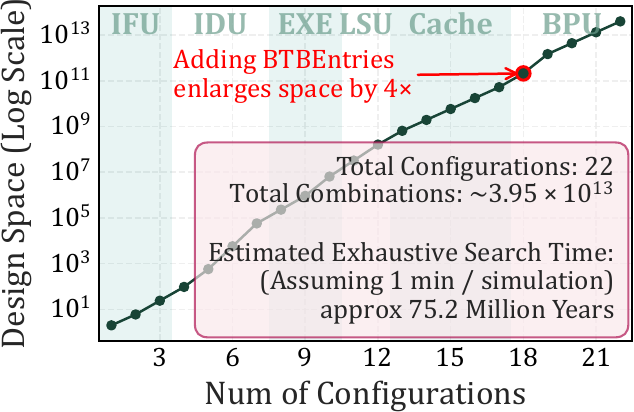}
\label{fig:ds_long}}
\hfil
\subfloat[]{\includegraphics[width=4.16cm]{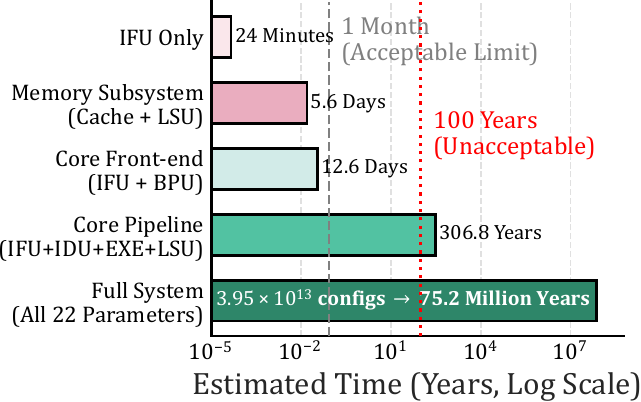}
\label{fig:ds_time}}
\vspace{-8pt}
\caption{(a) The combinatorial explosion of the microarchitecture configurations. (b) Exhaustive search time for CPU design space.}
\label{fig_ds}
\vspace{-8pt}
\end{figure}

\subsection{Microarchitecture DSE}

Microarchitecture DSE is constrained by an enormous combinatorial space. A modern out-of-order processor exposes many interacting parameters across the front-end, execution engine, and memory hierarchy. Even with only 22 basic configurations, the feasible design space reaches approximately $3.95 \times 10^{13}$ combinations, as illustrated in Figure \ref{fig:ds_long}. Moreover, obtaining reliable PPA metrics typically requires cycle-accurate simulation and power modeling, making each evaluation expensive \cite{profiling_gem5}. As shown in Figure \ref{fig:ds_time}, even under the optimistic assumption of one minute per simulation, exhaustively searching the entire design space would require approximately 75.2 million years. This exhaustive-search estimate implies that, under realistic project schedule constraints, the available evaluation budget is extremely scarce and every simulation opportunity is precious.

\begin{figure}[t]
\centering
\subfloat[]{\includegraphics[width=4.15cm]{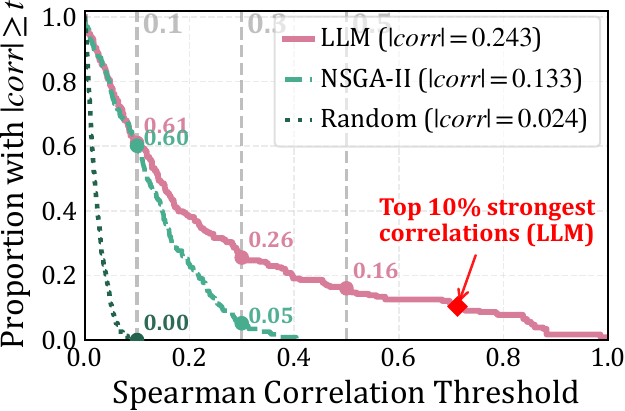}
\label{fig:config_corr}}
\hfil
\subfloat[]{\includegraphics[width=3.32cm]{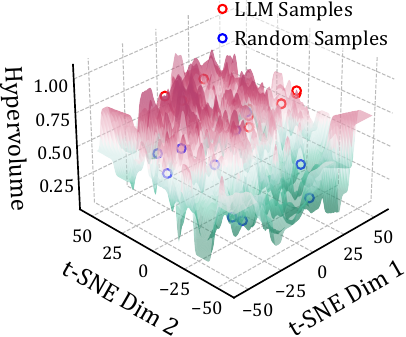}
\label{fig:ds_sample}}
\vspace{-8pt}
\caption{(a) The proportion of configuration pairs (e.g., \texttt{FetchWidth} and \texttt{RobEntries}) whose correlation strength exceeds a given threshold. A curve with a lower slope indicates a stronger correlation. (b) t-SNE visualization of microarchitecture design space.}
\label{fig_corr}
\vspace{-5pt}
\end{figure}

This challenge is further amplified by the ruggedness of the design space, as there is complex coupling between microarchitecture configurations. For example, enlarging the ROB capacity may yield performance gains only when the fetch width and the number of functional units are matched. Otherwise, it merely increases power and area overhead. Such coupling effects make the design space highly rugged, densely populated with local optima, where even minor perturbations in parameter configurations can cause unpredictable fluctuations in PPA outcomes (Figure \ref{fig:ds_sample}). To visualize this ruggedness, we reduce the configurations to a 2D space using t-SNE \cite{ArchExplorer}, with the PPA metrics and hypervolume \cite{hv} evaluated using the experimental setup detailed in Section \ref{sec:exp_setup}. As a result, unguided blind search is prone to getting trapped in local optima or wasting evaluations in low-quality regions without learning how microarchitectural components should be coordinated.

\vspace{-5pt}

\subsection{Challenges \& Motivations}

Given that coordination between microarchitecture configurations can introduce vast suboptimal solutions, blindly exploring the design space without professional knowledge is unwise. Incorporating prior knowledge into DSE has been recognized as a viable way to improve exploration efficiency. BOOM-Explorer \cite{boomexplorer} leverages active learning guided by the clustering features of configurations to accelerate exploration. RL-DSE \cite{rldse} models microarchitecture scaling graphs to inform configuration correlations, training an agent for component-wise parameter selection. 

\textbf{Challenge 1: Lack of explicit knowledge guidance.}
While these methods leverage prior knowledge to accelerate exploration, such experience is implicitly learned through task-specific models. This task-specific experience incurs high portability costs across architectures and cannot be directly queried or reused during DSE.

\begin{figure}[t]
\centering
\subfloat{\includegraphics[width=4cm]{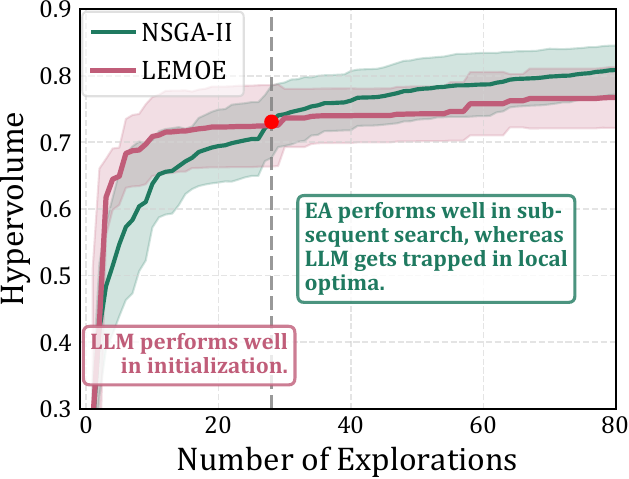}
\label{fig:hv_comparison}}
\hfil
\subfloat{\includegraphics[width=4cm]{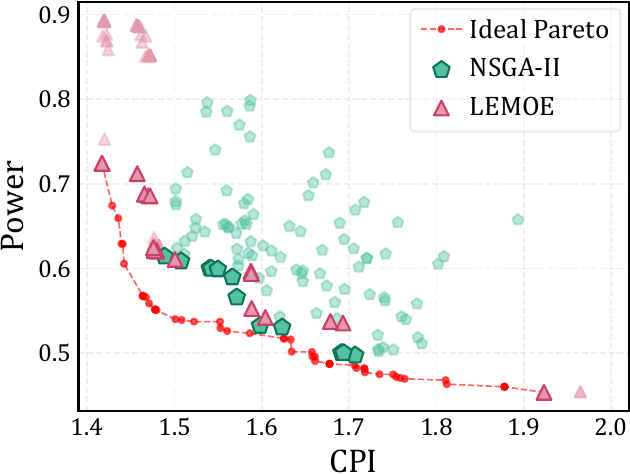}
\label{fig:ds_comparison}}
\vspace{-5pt}
\caption{Left: Comparison of the LLM-driven and EA-based DSE methods. Right: Search results in the CPI–Power space.}
\label{fig:lemoe}
\vspace{-5pt}
\end{figure}

\textbf{Why LLM.}
The successful application of an LLM to chip design and verification has inspired their adoption in hardware DSE \cite{chipmind, ChatCPU, chatsva, chathls, fixme, idse, chatmodel}. As shown in Figure \ref{fig:config_corr}, LLM-generated samples (DeepSeek-V3.2) exhibit stronger parameter correlations. This correlation pattern suggests that pretrained knowledge enables the LLM to capture the intrinsic configuration coupling. Moreover, when these samples are mapped into the objective space (Figure \ref{fig:ds_sample}), LLM-generated points identify high-quality Pareto regions (distributed in the upper half-space along the z-axis) compared to random sampling. These results highlight LLMs as promising knowledge-driven samplers. Recent LLM-driven DSE methods have also shown strong potential \cite{LEMOE, ChatA2}. As Figure \ref{fig:lemoe} (left) shows, they quickly approach a broad Pareto front, as reflected by the steep rise of the hypervolume curve. 

\textbf{Challenge 2: Ineffective utilization of iterative feedback.} 
Despite their high-quality sampling, existing LLM-driven DSE methods struggle to translate the knowledge advantages of LLMs into sustained optimization benefits. As exploration progresses, the lack of high-value iterative feedback to inform further search decisions often leads to a plateau in multi-objective optimization efficiency.

\textbf{Why evolutionary iteration.}
Without effective search state management and diversity preservation mechanisms, LLM-driven sampling tends to repeatedly exploit a few successful design patterns, leading to premature convergence in local optima. In contrast, as shown in Figure \ref{fig:lemoe}, traditional evolutionary algorithms maintain a stable optimization trajectory through population control, even though they start slower due to blind probabilistic operators. Therefore, combining the knowledge-guided generation by LLMs with the evolutionary framework offers a promising direction for high-efficiency sampling, sustained exploration, and global convergence.

\section{Design and Methodology}
\label{sec:Methodology}

\begin{figure*}[t]
	\centering 
 \includegraphics[width=\linewidth]{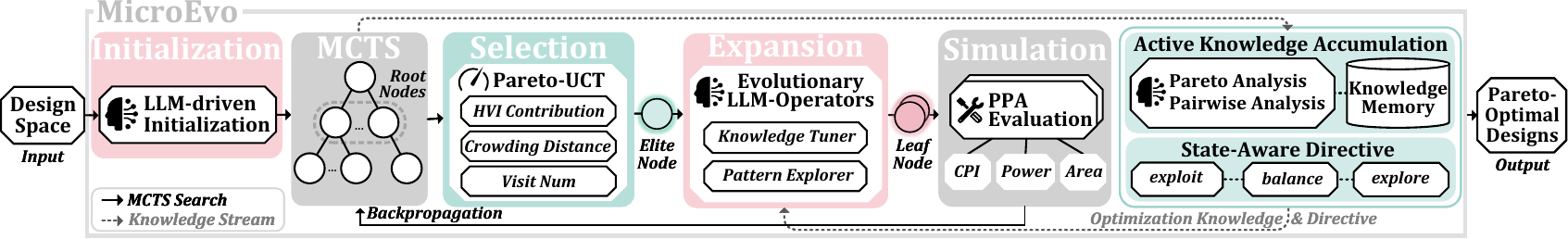} 
 \vspace{-15pt}
	\caption{Overview of MicroEvo. LLM first initializes root designs, Pareto-UCT then selects elite nodes in the MCTS tree \textit{(Section \ref{sec:pareto-uct})}, LLM-driven operators expand new nodes \textit{(Section \ref{sec:llm_expansion})}, and PPA evaluation updates the tree, while active knowledge accumulation \textit{(Section \ref{sec:knowledge})} and state-aware directives \textit{(Section \ref{sec:sad})} continuously provide optimization guidance  until the simulation budget is reached.}
	\label{fig:main}
 \vspace{-5pt}
\end{figure*}

This section presents the \textbf{MicroEvo} framework (Figure \ref{fig:main}). It formulates microarchitecture DSE as an evolutionary search process organized by Monte Carlo Tree Search (MCTS). The LLM first initializes a set of root designs targeting different optimization preferences. After selecting promising nodes using a multi-objective UCT criterion, MicroEvo applies LLM-driven evolutionary operators to expand the search tree, while learning from accumulated exploration knowledge and dynamically adapting the expansion directive. Each expanded node corresponds to a newly refined microarchitecture design, further steering the search toward the Pareto front.

\subsection{Evolutionary Decision Tree Search}
\label{sec:MCTS}

\begin{figure}[t]
	\centering 
 \includegraphics[width=7.5cm]{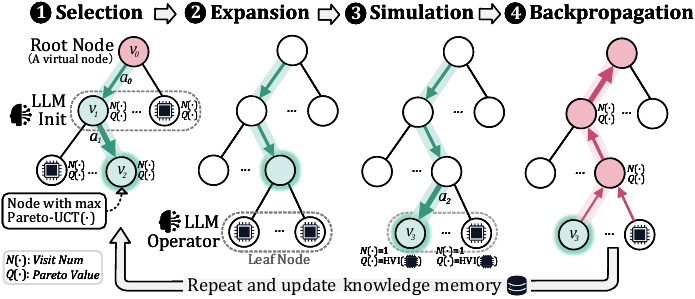} 
 \vspace{-5pt}
	\caption{Evolutionary Monte Carlo Tree Search (MCTS) procedure.}
	\label{fig:MCTS}
\vspace{-5pt}
\end{figure}

We instantiate MCTS as the search backbone of MicroEvo, where each tree expansion is implemented by LLM-driven mutation and crossover operators. In this formulation, MCTS provides structured selection and backpropagation \cite{mcts_review, Wavefront-MCTS}, while the LLM serves as the generator of new designs. In the search tree, each node (\textit{state} $v_i$) corresponds to a microarchitecture design with specific parameters. Each edge in Figure \ref{fig:MCTS} represents an \textit{action} $a_i$ that modifies the design, such as tuning configurations. The PPA simulation results of each optimized design serve as the \textit{reward function} $Q(v_i)$ for \textit{state} $v_i$. At each iteration, MCTS performs the following four stages.

\ding{182} \textit{Selection.}
Selection balances exploration and exploitation in MCTS. Starting from the root node, MicroEvo recursively selects the child with the highest upper confidence bound until reaching a leaf node. A standard UCT criterion is defined as
\begin{equation}
\label{eq:uct}
\mathrm{UCT}(v_i) = Q(v_i) + \lambda \sqrt{\frac{\ln N(v_p)}{N(v_i)}}
\end{equation}
where $Q(v_i)$ denotes the reward of child node $v_i$, $N(v_i)$ is its visit count, $v_p$ is its parent node, and $\lambda$ is the exploration coefficient. Unlike elite selection in genetic algorithms, UCT trades off current value and visit frequency, thereby allowing the search to revisit suboptimal yet potentially promising nodes. In our design, we further improve the UCT using multi-objective metrics (Section \ref{sec:pareto-uct}).

\ding{183} \textit{Expansion.}
Given the selected leaf node, MicroEvo invokes LLM-driven evolutionary operators to generate new candidate designs. Rather than relying on hand-crafted heuristics, these operators propose knowledge-guided modifications based on the current search state and historical exploration (Section~\ref{sec:llm_expansion}).

\ding{184} \textit{Simulation.}
In the simulation stage, the expanded design is evaluated using the target toolchain to obtain its PPA metrics.

\ding{185} \textit{Backpropagation.}
The evaluation result is propagated along the path to update values and visit counts of all encountered nodes.

After a predefined simulation count (budget) is exhausted, the MCTS tree contains a set of high-quality microarchitecture designs.

\subsection{Pareto-UCT}
\label{sec:pareto-uct}

Selecting which nodes to expand is a crucial step in MCTS \cite{mcts_review}. In Section \ref{sec:MCTS}, the standard UCT rule is defined by a single-objective reward. However, multi-objective microarchitecture DSE requires balancing competing PPA objectives, where search quality should be judged by its contribution to the global Pareto front rather than a single scalar. To address this mismatch, we propose \textbf{Pareto-UCT}, a selection criterion for global search.
For a child node $v_i$, its Pareto-UCT score is defined as
\vspace{-5pt}
\begin{equation}
\setlength{\abovedisplayskip}{1pt}
\setlength{\belowdisplayskip}{1pt}
\text{Pareto-UCT}(v_i) ={Q}_{\mathrm{HVI}}(v_i) + e\,{D}(v_i) + \lambda \sqrt{\frac{\ln N(v_p)}{N(v_i)}}
\label{eq:pareto_uct}
\end{equation}
where $v_p$ denotes the parent node of $v_i$, $N(v_p)$ and $N(v_i)$ are the visit counts of the parent and child nodes, respectively.

The first term ${Q}_{\mathrm{HVI}}(v_i)$ in Eq.~\eqref{eq:pareto_uct} quantifies the contribution of the current branch to the global Pareto front, measured by the hypervolume improvement (HVI) of node $v_i$:
\begin{equation}
\setlength{\abovedisplayskip}{1pt}
\setlength{\belowdisplayskip}{1pt}
\mathrm{HV}(\mathcal{P}) = \Lambda \left( \bigcup_{i=1}^{|\mathcal{P}|} [\pmb{f}(v_i), \pmb{r}] \right)
\label{eq:hv}
\end{equation}
\begin{equation}
{Q}_{\mathrm{HVI}}(v_i) = \mathrm{HV}\!\big(\mathcal{P} \cup \{\pmb{f}(v_i)\}\big) - \mathrm{HV}(\mathcal{P})
\label{eq:hvi}
\end{equation}
where $\mathcal{P}$ denotes the current non-dominated set, $\Lambda(\cdot)$ represents the hyper-rectangle bounded by the objective vector $\pmb{f}$ and the reference point $\pmb{r}$. Intuitively, ${Q}_{\mathrm{HVI}}(v_i)$ measures the marginal contribution of $v_i$ to expanding the dominated objective space volume.

The second term, ${D}(v_i)$ weighted by $e$, represents the crowding distance in objective space. This term prioritizes sparsely sampled regions, encouraging the search to cover under-explored parts of the Pareto front. Let $\mathcal{C}$ be the set of valid simulated children of $v_p$. We form a comparison set $\mathcal{S} = \mathcal{P} \cup \mathcal{C}$. For each node $v_i \in \mathcal{S}$, $D(v_i)$ accumulates the normalized distance between adjacent neighbors across all objective dimensions $\mathcal{M}$:
\begin{equation}
D(v_i) = \sum_{m=1}^\mathcal{M} \frac{f_m(v_i^{+,m}) - f_m(v_i^{-,m})}{f_m^{\max} - f_m^{\min}}
\label{eq:crowding_compact}
\end{equation}
where $\mathcal{M}=3$ corresponds to the PPA objectives, and $f_m^{\max}$ and $f_m^{\min}$ are the maximum and minimum values of objective $m$ in $\mathcal{S}$, respectively. The terms $v_i^{+,m}$ and $v_i^{-,m}$ denote the upper and lower neighbors of $v_i$ when the set $\mathcal{S}$ is sorted along objective $m$.

The third term ensures that infrequently visited and suboptimal yet promising nodes still have a chance to be explored:
\begin{equation}
\setlength{\abovedisplayskip}{1pt}
\setlength{\belowdisplayskip}{1pt}
\lambda = \lambda_0 \sqrt{\frac{T - t}{T}}
\label{eq:lambda_decay}
\end{equation}
where $\lambda$ is a dynamically decayed exploration coefficient, with $\lambda_0$ as the initial weight, $t$ the current iteration, and $T$ the total search budget. This schedule gradually shifts the search from early exploration to later exploitation.

Using this criterion, MicroEvo selects an elite leaf node for subsequent expansion. Compared with standard UCT, Pareto-UCT preserves the exploration--exploitation trade-off while incorporating Pareto front quality and distribution diversity, making it more suitable for multi-objective microarchitecture DSE.

\begin{figure*}[t]
	\centering 
 \includegraphics[width=\linewidth]{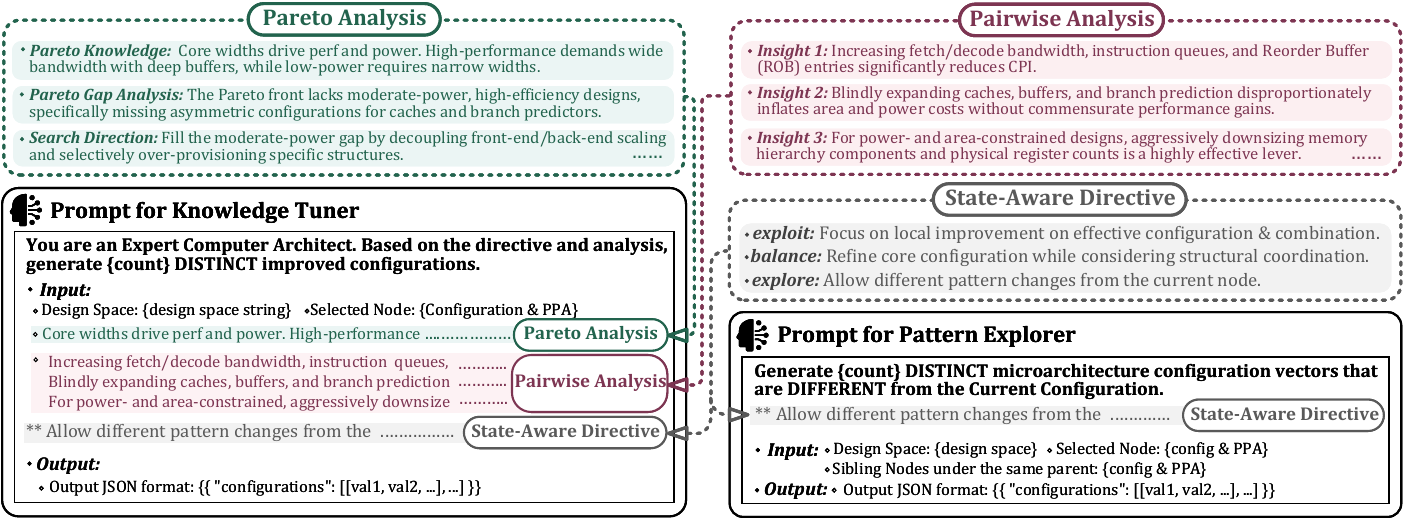} 
 \vspace{-16pt}
	\caption{Prompt integration for LLM-driven expansion.}
	\label{fig:prompt}
\vspace{-10pt}
\end{figure*}

\vspace{-5pt}

\subsection{LLM-Driven Initialization \& Expansion}
\label{sec:llm_expansion}

Unlike traditional MCTS that builds from a single node, MicroEvo begins from a virtual root where the LLM generates an initial set of microarchitecture designs. To provide a broader spectrum of search directions early in DSE, we instruct the LLM to generate distinct parameter combinations targeting specific PPA trade-offs: maximum performance, maximum efficiency, balanced, performance-leaning, and power/area-leaning. The LLM selects configurations from a user-defined design space, leveraging its internalized knowledge of microarchitectural correlations to produce effective designs. To handle potential LLM hallucinations, any generated parameter outside the design space is mapped to the nearest valid value.

Once the initial set is evaluated and the MCTS tree is established, Pareto-UCT continuously selects elite nodes for expansion. To better inform subsequent search decisions, MicroEvo integrates PPA feedback and expert knowledge into structured prompts that drive the LLM to propose new children, as shown in Figure \ref{fig:prompt}. Inspired by the mutation and crossover mechanisms in evolutionary algorithms \cite{CogMCTS, HiFo-Prompt}, we design two distinct LLM-driven expansion operators: the \textbf{knowledge tuner} and the \textbf{pattern explorer}.

The \textit{knowledge tuner} leverages global insights (Section \ref{sec:knowledge}) and parent node PPA results for local improvements. Based on the accumulated optimization knowledge, it avoids inefficient combinations. For example, if historical experience indicates that scaling the reorder buffer (ROB) requires increasing load/store queues to prevent stalls, the LLM adjusts these coupled parameters accordingly. Moreover, by identifying objectives lagging behind the Pareto front (e.g., power efficiency), the LLM translates this gap into targeted refinements, such as downsizing caches or issue widths.

The \textit{pattern explorer} goes beyond random parameter mutation to identify higher-level optimization patterns. Given sibling nodes that share the same parent, the LLM analyzes their configurations and PPAs to propose novel solutions relative to the current set of explored nodes. For example, recognizing a ``high fetch/issue width + large cache capacity'' pattern suited for complex workloads, it may propose an alternative optimization regime. In this way, the mutation operates at the level of architectural design intent rather than superficial parameter perturbation.

\begin{algorithm}[t]
\small
\caption{Active Knowledge Accumulation with Knowledge-Guided LLM Expansion}
\label{alg:aka}
\begin{algorithmic}[1]
\Require selected node $v$, sibling set $\mathcal{S}(v)$, Pareto set $\mathcal{P}$, insight memory $\Omega$
\Ensure child node $\mathcal{C}_{\mathrm{new}}$, updated Pareto set $\mathcal{P}^*$, updated memory $\Omega^*$

\State $\textit{report} \gets \text{ParetoAnalysis}(v,\mathcal{P})$ \hfill \footnotesize $\triangleright$ returns Pareto knowledge

\ForAll{$k \in \Omega$}
    \State compute utility
    $U(k) = \widetilde{Q}_{\mathrm{HVI}}(k) - \omega \log(N(k)+1)$
\EndFor
\State $\mathcal{K} \gets \text{Top-K}(\Omega, U)$ \Comment{retrieve the most useful insights}

\State $\mathcal{C}_{r} \gets \text{KnowledgeTuner}(v,\textit{report},\mathcal{K})$ \Comment{guided by Pareto and pairwise analysis}
\State $\mathcal{C}_{p} \gets \text{PatternExplorer}(v,\mathcal{S}(v))$ \Comment{exploration using sibling context}
\State $\mathcal{C}_{\mathrm{new}} \gets \mathcal{C}_{r} \cup \mathcal{C}_{p}$

\ForAll{$c \in \mathcal{C}_{\mathrm{new}}$}
    \State $\mathrm{CPI}, \mathrm{Power}, \mathrm{Area} \gets \text{Evaluate}(c)$
    \State $\mathrm{HVI}(c) \gets \text{NonDominatedSort}(c,\mathcal{P})$
        \If{$\mathrm{HVI}(c) > 0.01$ \textbf{or} $\exists j,\ (f_j(v)-f_j(c))/f_j(v) > 20\%$}
        \State $k_{\mathrm{new}} \gets \text{PairwiseAnalysis}(v,c)$ \Comment{extract from parent-child PPA change}
        \If{$\text{NotDuplicate}(k_{\mathrm{new}},\Omega)$} \Comment{similarity check via embedding}
            \State $\Omega^{*} \gets \Omega \cup \{k_{\mathrm{new}}\}$
        \EndIf
    \EndIf
\EndFor

\State update $N(k)$, Pareto-optimal design set, and cumulative HVI gain

\State \Return $\mathcal{C}_{\mathrm{new}}, \mathcal{P}^{*}, \Omega^{*}$
\end{algorithmic}
\end{algorithm}

\vspace{-10pt}

\subsection{Active Knowledge Accumulation}
\label{sec:knowledge}

To extract practical experience and inform LLM decision-making during expansion, we introduce an \textbf{active knowledge accumulation (AKA)} mechanism together with a \textit{knowledge memory} to manage experience extraction and application. Whenever a child node is expanded, the LLM reflects on the optimization trajectory and summarizes reusable insights (Algorithm \ref{alg:aka}). These generated insights are grounded in the PPA changes and are directly relevant to the ongoing DSE process. Therefore, AKA transforms process-level experiences into reusable knowledge across two levels.

\textit{Pareto Analysis.}
By performing non-dominated sorting over the explored designs, AKA identifies the current Pareto-optimal microarchitectures and analyzes the coverage of the explored front. This analysis allows the LLM to summarize multi-objective optimization insights across diverse design preferences. It learns to construct parameter combinations that approximate the broad Pareto front and extracts global optimization patterns. Moreover, it diagnoses which regions of the objective space remain under-explored and suggests the next search direction. To maintain solution diversity, this guidance avoids prescribing exact configuration edits.

\textit{Pairwise Analysis.} AKA compares each expanded child with its parent, correlating parameter adjustments with resulting PPA variations to translate raw simulation feedback into actionable tuning rules. For instance, if reducing L2 cache associativity decreases power consumption with minimal performance loss, the LLM formulates a guideline prioritizing this case for power-constrained designs. This process extracts fine-grained procedural experience on beneficial configuration updates. To avoid accumulating trivial knowledge, extraction is triggered only when the child delivers a significant HVI gain or clear improvement in at least one objective. To ensure that the extracted insights are robust to potential absolute simulation errors from analytical tools (e.g., McPAT), the LLM is instructed to avoid generating specific quantitative mappings.

Since not all extracted insights are equally valuable, AKA employs a \textbf{knowledge memory} with redundancy-aware insertion and utility-aware retrieval. Each memory item stores one reusable insight with its historical application records. New insights are semantically filtered before insertion to remove redundant items. When insight $k$ is used, AKA records the expanded child node $v_t$ and the obtained HVI. The memory module then scores each stored insight and retrieves the highest-scoring items to guide the next expansion. The utility score $U$ of insight $k$ is defined as
\begin{equation}
\setlength{\abovedisplayskip}{1pt}
\setlength{\belowdisplayskip}{1pt}
U(k) = \underbrace{{\tilde{Q}}_{\mathrm{HVI}}(v_i \mid k)}_{\textit{Pareto Reward}} - \underbrace{\omega \log(N(k)+1)}_{\textit{Visit Penalty}}
\label{eq:memory_utility}
\end{equation}
\begin{equation}
{\tilde{Q}}_{\mathrm{HVI}}(v_i \mid k) = \frac{Q_{\mathrm{HVI}}(v_i \mid k) -  Q^{min}_{\mathrm{HVI}}(v_i)} {Q^{max}_{\mathrm{HVI}}(v_i) - Q^{min}_{\mathrm{HVI}}(v_i) +\epsilon}
\end{equation}
where $N(k)$ is the usage count of $k$, $\omega$ controls the penalty strength, and $\epsilon$ is a small constant to prevent division by zero. The utility score balances two factors: \textit{Pareto Reward} uses normalized HVI gain to reflect the contribution to the Pareto front, and \textit{Visit Penalty} suppresses frequently reused insights to encourage visiting newer knowledge. This memory mechanism prevents a few rules from dominating the search while prioritizing consistently effective and recently successful insights, ensuring that the retrieved knowledge remains reusable and adaptive as the search trajectory evolves.

\begin{figure}[t]
	\centering 
 \includegraphics[width=\linewidth]{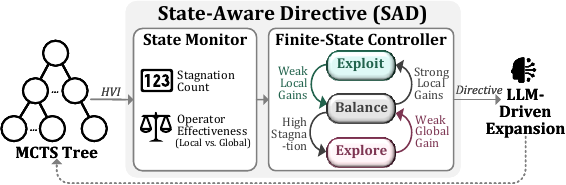} 
 \vspace{-14pt}
	\caption{State-Aware Directive (SAD) workflow}
	\label{fig:sad}
\end{figure}

\vspace{-8pt}

\subsection{State-Aware Directive}
\label{sec:sad}

A fixed prompting strategy is often insufficient for LLM-driven microarchitecture DSE \cite{ChatArch}. When the search progresses well, the framework should favor minor refinements around promising designs. In contrast, when the search stalls, repeated local modifications waste the simulation budget. To address this issue, we introduce a \textbf{state-aware directive (SAD)} mechanism, which monitors the search progress and dynamically adjusts the directive given to the LLM during node expansion (Figure \ref{fig:sad}).

SAD operates as a finite-state controller with three modes: \textit{exploit}, \textit{balance}, and \textit{explore}. It tracks two runtime signals. The first is \texttt{stagnation\_count}, which records the number of consecutive evaluations without meaningful HVI (1\% of the total HV) and resets upon any significant HVI gain. The second is the recent effectiveness of expansion operators, indicating whether the current search is benefiting more from local refinement or from broader structural changes. Based on these signals, SAD updates the search mode in real time. If \texttt{stagnation\_count} exceeds a predefined threshold, the search is considered trapped locally, and the controller switches to \textit{explore}. If stagnation is low and recent local refinements consistently improve the Pareto front, it switches to \textit{exploit}. In all other cases, the controller stays in \textit{balance}.

\begin{table*}[t]
\centering
\caption{Comparison of different DSE methods under two search budgets. MicroEvo is evaluated with DeepSeek-V3.2 and Gemini-3-pro.}
\vspace{-10pt}
\label{tab:overall_results}
\setlength{\tabcolsep}{1.5pt}
\renewcommand{\arraystretch}{1.2}
\resizebox{\linewidth}{!}{
\begin{tabular}{l|cc|cc|cc|cc|cc|cc}
\noalign{\hrule height 1.0pt}
\multirow{2}{*}{Method} 
& \multicolumn{6}{c|}{Evaluation Budget = 20}
& \multicolumn{6}{c}{Evaluation Budget = 45} \\
\cline{2-13}
& HV Mean $\uparrow$ {\small $\pm$ Std} & Imp. $\uparrow$ & ADRS Mean $\downarrow$ {\small $\pm$ Std} & Imp. $\uparrow$ & IPC/Power $\uparrow$ & IPC/Area $\uparrow$
& HV Mean $\uparrow$ {\small $\pm$ Std} & Imp. $\uparrow$ & ADRS Mean $\downarrow$ {\small $\pm$ Std} & Imp. $\uparrow$ & IPC/Power $\uparrow$ & IPC/Area $\uparrow$ \\
\noalign{\hrule height 1.0pt}
NSGA-II        
& 0.5826 {\small $\pm$ 0.0439} & 0.00\%  & 0.1666 {\small $\pm$ 0.0213} & 0.00\%   & 1.1823 & 0.0487
& 0.663 {\small $\pm$ 0.022}  & 0.00\%  & 0.123 {\small $\pm$ 0.017}  & 0.00\%   & 1.1823 & 0.0524 \\
MOTPE \cite{motpe}        
& 0.6058 {\small $\pm$ 0.0261} & +3.99\% & 0.1454 {\small $\pm$ 0.0171} & +12.76\% & 1.1762 & 0.0510
& 0.698 {\small $\pm$ 0.026}  & +5.42\% & 0.100 {\small $\pm$ 0.011}  & +18.95\% & 1.2053 & 0.0513 \\
LEMOE \cite{LEMOE}       
& 0.6746 {\small $\pm$ 0.0268} & +15.80\%& 0.1315 {\small $\pm$ 0.0249} & +21.11\% & 1.1568 & 0.0498
& 0.679 {\small $\pm$ 0.024}  & +2.53\% & 0.128 {\small $\pm$ 0.021}  & -3.87\%  & 1.1568 & 0.0498 \\
Boom-Explorer \cite{boomexplorer}
& 0.6203 {\small $\pm$ 0.0528} & +6.48\% & 0.1562 {\small $\pm$ 0.0194} & +6.29\%  & 1.1854 & 0.0492
& 0.709 {\small $\pm$ 0.042}  & +6.97\% & 0.100 {\small $\pm$ 0.010}  & +19.12\% & \underline{1.2276} & \textbf{0.0537} \\
RL-DSE \cite{rldse}
& 0.6340 {\small $\pm$ 0.0577} & +8.82\% & 0.1444 {\small $\pm$ 0.0259} & +13.33\% & 1.1604 & \underline{0.0526}
& 0.711 {\small $\pm$ 0.012}  & +7.36\% & 0.105 {\small $\pm$ 0.008}  & +14.46\% & 1.1942 & 0.0526 \\
\hline
\rowcolor[HTML]{E3F5EE} 
\textbf{MicroEvo (DeepSeek)}   
& \underline{0.7822 {\small $\pm$ 0.0179}} & \underline{+34.27\%} & \underline{0.0772 {\small $\pm$ 0.0112}} & \underline{+53.65\%} & \textbf{1.2156} & 0.0521
& \underline{0.817 {\small $\pm$ 0.007}}  & \underline{+23.35\%} & \underline{0.056 {\small $\pm$ 0.005}}  & \underline{+54.68\%} & 1.2156 & 0.0523 \\
\rowcolor[HTML]{E3F5EE} 
\textbf{MicroEvo (Gemini)}   
& \textbf{0.7936} {\small $\pm$ \textbf{0.0059}} & \textbf{+36.22\%} & \textbf{0.0716} {\small $\pm$ \textbf{0.0063}} & \textbf{+57.06\%} & \underline{1.1951} & \textbf{0.0527}
& \textbf{0.821} {\small $\pm$ \textbf{0.005}}  & \textbf{+23.94\%} & \textbf{0.056} {\small $\pm$ \textbf{0.003}}  & \textbf{+54.80\%} & \textbf{1.2360} & \underline{0.0530} \\
\noalign{\hrule height 1.0pt}
\end{tabular}}
\\[1 pt]
\raggedright 
{\scriptsize 
\renewcommand{\baselinestretch}{1.2}
\noindent *~ \textbf{Bold} and \underline{underline} indicate the best and second-best results, respectively. HV Imp. is calculated as the relative improvement compared to the baseline DSE method, whereas ADRS is the relative reduction. \par}
\vspace{-8pt}
\end{table*}

\begin{table}[htbp]
\centering
\caption{Microarchitecture design space configuration}
\vspace{-10pt}
\label{tab:design_space}
\small
\renewcommand{\arraystretch}{0.98}
\setlength{\tabcolsep}{2pt} 
\resizebox{\columnwidth}{!}{
\begin{tabular}{|l|l||l|l|}
\hline
\textbf{Components} & \textbf{Candidate Values} & \textbf{Components} & \textbf{Candidate Values} \\ \hline \hline

\multicolumn{2}{|c||}{\textbf{Alpha21264}} & \multicolumn{2}{c|}{\textbf{XiangShan (continued)}} \\ \hline

\texttt{FetchWidth}      & 4, 8 & \texttt{phyregReleaseWidthInt} & 1, 2, 4, 6, 8 \\ \hline

\texttt{FetchBufferSize} & 16, 32, 48, 64 & \texttt{phyregReleaseWidthFp}  & 1, 2, 4, 6, 8 \\ \hline

\texttt{FetchQueueSize}  & 16, 32, 48, 64 & \texttt{phyregReleaseWidthVec} & 1, 2, 4 \\ \hline

\texttt{DecodeWidth}     & 2, 4, 6, 8 & \texttt{ALUQueueSize}         & 12:36:6 \\ \hline

\texttt{ROBSize}             & 32:192:32 & \texttt{FVUQueueSize}         & 12:36:6 \\ \hline

\texttt{IntPhyRegister}          & 32:320:32 & \texttt{MDUQueueSize}         & 12:36:6 \\ \hline

\texttt{FpPhyRegister}           & 32:320:32 & \texttt{ALUQueueEnqSize}      & 1, 2, 3 \\ \hline

\texttt{DispatchWidth}   & 2, 4, 6, 8 & \texttt{FVUQueueEnqSize}      & 1, 2, 3 \\ \hline

\texttt{IssueWidth}      & 2, 4, 6, 8 & \texttt{MDUQueueEnqSize}      & 1, 2, 3 \\ \hline

\texttt{IQEntries}              & 16:96:8 & \texttt{LSQSize}         & 12:36:6 \\ \hline

\texttt{LDQEntries}             & 16, 24, 32, 40, 48 & \texttt{LSQEnqSize}      & 1, 2, 3 \\ \hline

\texttt{STQEntries}             & 16, 24, 32, 40, 48 & \texttt{LoadRequestQueueSize}& 16, 32, 64 \\ \hline

\texttt{TLBSize}         & 64, 96, 128, 160 & \texttt{TempStoreQueueSize}& 16, 32, 64 \\ \hline

\texttt{ICacheSize}      & 16\,kB, 32\,kB, 64\,kB & \texttt{StoreMergeBuffer}& 8, 16, 32 \\ \hline

\texttt{ICacheAssociativity}     & 2, 4, 8 & \texttt{RARQSize}        & 24:128:8 \\ \hline

\texttt{DCacheSize}      & 16\,kB, 32\,kB, 64\,kB & \texttt{RAWQSize}        & 24:96:8 \\ \hline

\texttt{DCacheAssociativity}     & 2, 4, 8 & \texttt{LFSTSize}        & 8, 16, 32 \\ \hline

\texttt{BTBEntries}      & 1024, 2048, 4096, 8192 & \texttt{LFSTEntrySize}   & 1, 2, 3, 4 \\ \hline

\texttt{RASSize}         & 16:40:4 & \texttt{SSITSize}        & 8, 16, 32 \\ \hline

\texttt{LocalPredictorSize}  & 512, 1024, 2048 & \texttt{cacheLoadPorts}  & 1:8:1 \\ \hline

\texttt{GlobalPredictorSize} & 2048, 4096, 8192 & \texttt{cacheStorePorts} & 1:8:1 \\ \hline

\texttt{ChoicePredictorSize} & 2048, 4096, 8192 & \texttt{TLBSize}         & 32, 64, 128, 256 \\ \hline

\multicolumn{2}{|c||}{\textbf{Design Space: $\approx \mathbf{3.95 \times 10^{13}}$}} & \texttt{ICacheSize}      & 16\,kB, 32\,kB, 64\,kB \\ 

\hline

\multicolumn{2}{|c||}{\textbf{XiangShan}} & \texttt{ICacheAssociativity} & 4, 8 \\ \hline

\texttt{FetchWidth}      & 4, 8, 16 & \texttt{L1DcacheSize}    & 16\,kB, 32\,kB, 64\,kB \\ \hline

\texttt{FetchQueueSize}  & 16:96:16 & \texttt{DCacheAssociativity} & 4, 8 \\ \hline

\texttt{DecodeWidth}     & 4:12:2 & \texttt{MSHRSize}        & 8, 16, 32 \\ \hline

\texttt{RenameWidth}     & 4:12:2 & \texttt{uBTBSize}        & 512, 1024, 2048 \\ \hline

\texttt{ROBSize}         & 160:640:40 & \texttt{L1BTBSize}       & 2048, 4096, 8192 \\ \hline

\texttt{IntPregsNum}  & 160:640:20 & \texttt{L2BTBSize}       & 4096, 8192, 16384, 32768 \\ \hline

\texttt{FloatPregsNum}& 96:316:20 & \texttt{RASSize}         & 16, 32, 64 \\ \hline

\texttt{VectorPregsNum}& 48:96:8 & \multicolumn{2}{c|}{\textbf{Design Space: $\approx \mathbf{3.41 \times 10^{26}}$}} \\ \hline

\texttt{PredicateVectorPregsNum} & 96:368:16 & \multicolumn{2}{c|}{} \\ \hline

\end{tabular}}
\end{table}

The selected mode provides language directives to the LLM operator for expansion. In the \textit{exploit} mode, SAD injects a directive to the prompt to make local adjustments, preserving the main combination pattern. In the \textit{balance} mode, the LLM is guided to introduce moderate changes that consider configuration dependencies. In the \textit{explore} mode, the directive discourages minor edits and encourages larger structural changes targeting under-explored Pareto regions. When the current branch shows limited recent progress, this strategy helps the search move away from local design patterns. By introducing SAD, MicroEvo avoids relying on a static expansion policy. Instead, it dynamically adapts LLM-generated modifications according to the observed search state, allowing DSE to efficiently alternate between local refinement and broader exploration.


\section{Evaluations}
\label{sec:Evaluation}

\subsection{Experimental Setup}
\label{sec:exp_setup}

\textbf{Design space.}
We employ GEM5 \cite{gem5} for cycle-accurate simulation and McPAT \cite{mcpat} for power and area modeling. The primary evaluated processor is an out-of-order RISC-V core with a two-level cache hierarchy and 16 GB DRAM, with its design space detailed in Table \ref{tab:design_space}. For each design point, the PPA metrics are reported as the geometric mean across the RISC-V benchmark suite \cite{riscv-test}. To further validate the generalization and transferability of our framework, we also evaluate the XiangShan Kunminghu processor \cite{xiangshan} via XS-GEM5 \cite{xsgem5}. This processor is substantially more complex, featuring a TAGE branch predictor and multiple hardware prefetchers.

\textbf{Baselines.}
We compare our method with NSGA-II, MOTPE \cite{motpe}, Boom-Explorer (Bayesian optimization) \cite{boomexplorer}, RL-DSE (reinforcement learning) \cite{rldse}, and LEMOE (LLM-driven) \cite{LEMOE}. For RL-DSE, we allocate the evaluation budgets to obtain uniformly sampled preference vectors to learn the convex coverage set. Among prior LLM-driven DSE methods \cite{ChatArch, ChatDSE}, we select LEMOE as the most representative recently published method for this problem and reproduce it with DeepSeek-V3.2. All experiments are conducted on an Intel Xeon Platinum 8480+ server and averaged over five independent trials.

\textbf{DSE performance.}
We compare these DSE methods using hypervolume (HV) and average distance to reference set (ADRS). For HV, the reference point is defined by taking the maximum value along each objective. For ADRS, we construct the reference Pareto front $\mathcal{P}^{*}$ by combining 5{,}000 randomly sampled design points with all solutions explored by the evaluated methods:
\begin{equation}
\setlength{\abovedisplayskip}{2pt}
\setlength{\belowdisplayskip}{2pt}
\mathrm{ADRS}(\mathcal{P}, \mathcal{P}^{*}) = \frac{1}{|P^{*}|} \sum_{x^{*} \in \mathcal{P}^{*}} \min_{x \in \mathcal{P}} \delta(x, x^{*})
\end{equation}
\begin{equation}
\setlength{\abovedisplayskip}{2pt}
\setlength{\belowdisplayskip}{2pt}
\delta(x, x^{*}) = \max_{m \in \mathcal{M}} \left\{0, \frac{f_m(x) - f_m(x^{*})}{f_m(x^{*})} \right\}
\end{equation}
Here, $\mathcal{M}$ denotes the set of objectives. A lower ADRS indicates that the explored Pareto set is closer to the reference front.

To evaluate sampling efficiency, we measure the number of evaluations $t$ required for each method to reach a target threshold $\tau$:
\begin{equation}
\setlength{\abovedisplayskip}{2pt}
\setlength{\belowdisplayskip}{2pt}
N_{\tau} = \min \{ t \mid \mathrm{HV}(t) \geq \tau \},
\end{equation}
where $\mathrm{HV}(t)$ denotes the best HV achieved by a DSE method after $t$ evaluations. A smaller $N_{\tau}$ indicates higher search efficiency.

\textbf{Hyperparameter settings.}
The MicroEvo process runs for 10 iterations, starting from five initially sampled points (for a total of 45 simulations). During each expansion step, the two LLM operators each generate two new candidate designs. At each iteration, three insights are injected into the search. Newly generated insights are encoded via MiniLM embeddings, and semantic redundancies are filtered using a similarity threshold of 0.75. For Pareto-UCT, the parameter $e$ is set to 0.05 and $\lambda_0$ is set to 1.414. In Active Knowledge Accumulation (AKA), the penalty coefficient for repeated access to the same insight is 0.3.

\subsection{LLM-Driven MicroEvo Versus Prior Arts}

\begin{figure}[t]
	\centering 
 \includegraphics[width=\linewidth]{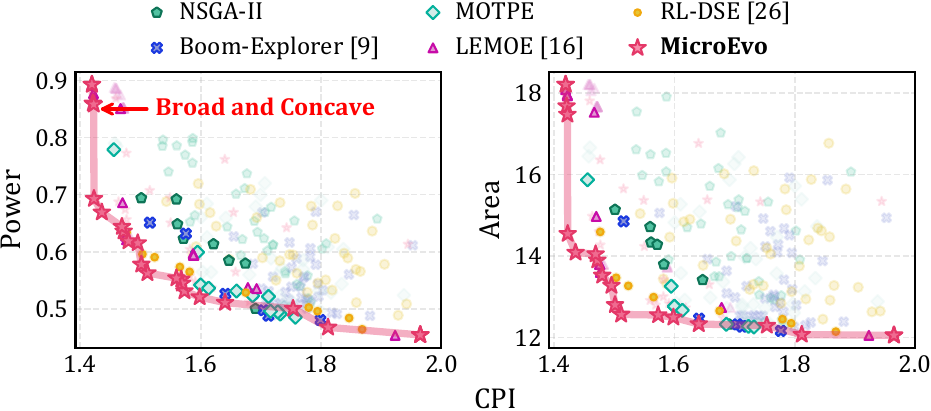} 
 \vspace{-23pt}
	\caption{Pareto front comparison of different DSE methods in the CPI–Power and CPI–Area spaces.}
	\label{fig:pareto_front}
    \vspace{-8pt}
\end{figure}

\textbf{Pareto quality comparison.} Table \ref{tab:overall_results} summarizes the optimization quality of the evaluated DSE methods. MicroEvo with Gemini-3-pro improves HV by 23.9\% over NSGA-II, 17.0\% over Boom-Explorer, 16.6\% over RL-DSE, and 21.4\% over LEMOE. Figure \ref{fig:pareto_front} further shows that MicroEvo achieves better Pareto front coverage and stays closer to the reference front under the same evaluation count. When the budget is reduced to only 20 evaluations (4 iterations for MicroEvo), it still identifies a broader set of Pareto-optimal solutions. This result is important for practical processor development, where the available simulation budget is typically very limited.

\begin{figure}[t]
	\centering 
 \includegraphics[width=\linewidth]{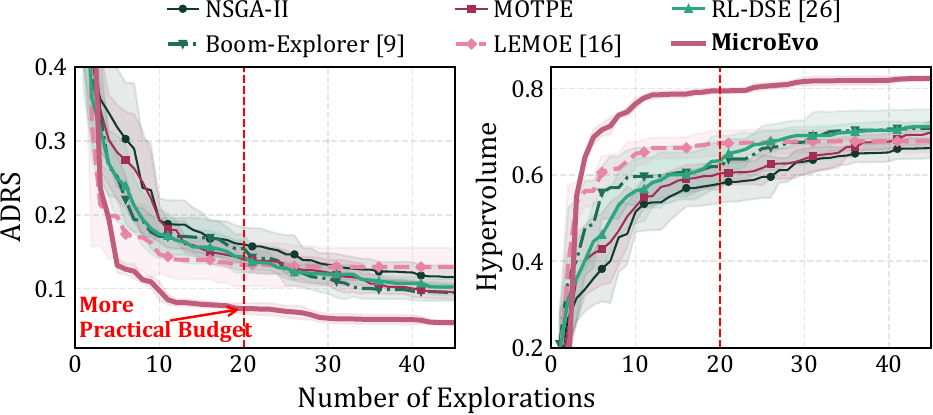} 
 \vspace{-24pt}
	\caption{Hypervolume and ADRS comparison of different DSE methods versus number of explorations: the red dashed line marks a more practical evaluation budget of 20 trials.}
    \vspace{-5pt}
	\label{fig:hv_adrs_curve}
\end{figure}

Traditional multi-objective optimization algorithms, such as NSGA-II and MOTPE, lack architectural domain knowledge and struggle to converge rapidly in a highly rugged design space. While Boom-Explorer benefits from Bayesian optimization (BO), its later search quality is constrained by Gaussian process prediction errors, leading to suboptimal results. RL-DSE requires iterative policy training before it can provide useful actions, making DSE inefficient under tight budgets. Furthermore, its learned knowledge is implicitly encoded in model parameters during training, rather than explicitly extracted and reused to guide exploration. LEMOE uses the same backbone LLM as MicroEvo, but still exhibits a clear performance gap. As illustrated in Figure \ref{fig:hv_adrs_curve}, this gap primarily emerges during the later search stages. Although LEMOE follows a BO-style iterative process, it makes limited use of PPA metrics to inform subsequent decisions. Moreover, the bias introduced by LLM-based PPA predictions further degrades candidate selection \cite{ChatDSE}. In contrast, MicroEvo continuously accumulates useful microarchitecture insights via AKA, while State-Aware Directive (SAD) dynamically adjusts search directives based on exploration status. Consequently, MicroEvo achieves a balance between exploration and exploitation.

\begin{figure}[t]
	\centering 
 \includegraphics[width=\linewidth]{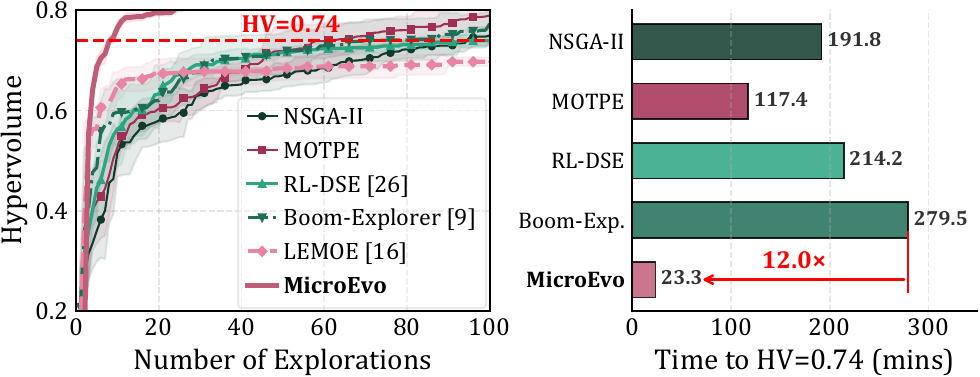} 
 \vspace{-20pt}
	\caption{Convergence efficiency and time-to-target comparison.}
    \vspace{-5pt}
	\label{fig:efficiency}
\end{figure}

\textbf{Comparison of sample efficiency.} To further evaluate search efficiency, we record the number of evaluations required to reach the target HV threshold ($HV=0.74$) in Figure \ref{fig:efficiency}. The results show that MicroEvo (Gemini) requires significantly fewer evaluations than the baselines, improving search efficiency by \textbf{10.6$\bm\times$} compared to NSGA-II, and consumes only \textbf{8.3\%} of Boom-Explorer's execution time to reach this target. In the early DSE stage, the synergy of Pareto analysis and LLM-pretrained knowledge of microarchitectural component interdependencies enables MicroEvo to rapidly approximate the Pareto front. This synergy ensures that each trial maintains structural validity against configuration coordinations and preserves a global view of the design space. In the later stage, accumulated knowledge and SAD allow the framework to dynamically adapt search directives based on the current state, helping the search escape local optima. Meanwhile, Pareto-UCT ensures that exploration focuses on the most promising nodes. It exploits the architectural patterns of the current best solutions and extracts valuable features from suboptimal designs, translating them into effective directions for subsequent exploration.

\begin{figure}[t]
	\centering 
\includegraphics[width=\linewidth]{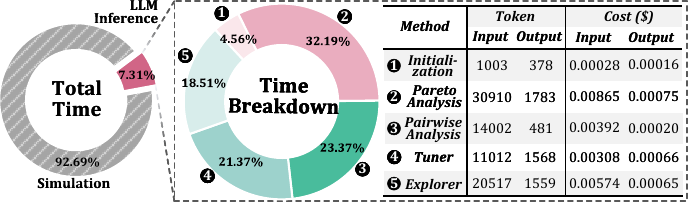} 
 \vspace{-15pt}
	\caption{Token usage, cost, and total runtime breakdown of MicroEvo
(DeepSeek-V3.2): 7.3\% LLM inference vs. 92.7\% simulation.}
	\label{fig:consumption}
    \vspace{-5pt}
\end{figure}

\begin{figure}[t]
\includegraphics[width=\linewidth]{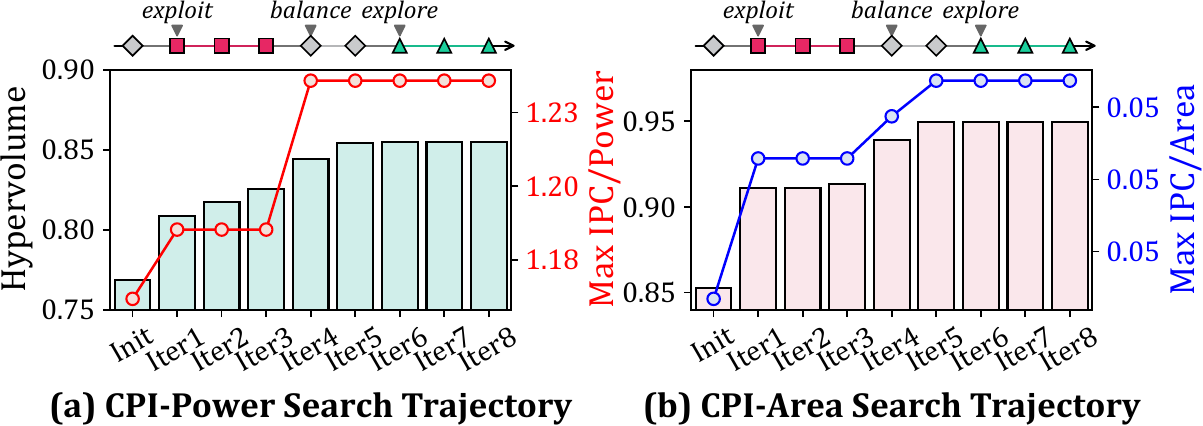} 
 \vspace{-20pt}
	\caption{MCTS search progress over iterations.}
	\label{fig:trajectory}
    \vspace{-5pt}
\end{figure}

\textbf{Practical overhead and search trajectory.} To quantify the LLM overhead in MicroEvo, Figure \ref{fig:consumption} reports the response time, token consumption, and API cost of each LLM-driven stage when using DeepSeek-V3.2. In one complete MicroEvo search, including initial sampling stage and ten rounds of iterative expansion, the total LLM invocation time is only 5.85 minutes, with an API cost of approximately \$0.024 (34.52 minutes of LLM invocation and \$1.32 API cost for Gemini-3-pro). LLM inference accounts for only 7.31\% of the overall runtime, indicating that MicroEvo remains practical and scalable for microarchitecture DSE. Figure \ref{fig:trajectory} shows that MicroEvo steadily improves search quality over iterations and reaches strong optimization performance with few iterations. The switching among the \textit{exploit}, \textit{balance}, and \textit{explore} modes further suggests that the state-aware directive can adapt the search behavior according to the current progress, enabling effective coordination between local refinement and broader exploration.

\subsection{Ablation Study}

\textbf{Learning from iterative search.} To validate the necessity of each MicroEvo component for efficient multi-objective optimization, we conduct an ablation study (Table \ref{tab:ablation_results}). Removing both active knowledge accumulation and the search directive (\textit{w/o AKA, SAD}) significantly degrades performance. This result indicates that simply combining LLMs with tree search is not sufficient for microarchitecture DSE. Without actionable knowledge support and search-state control, the framework fails to effectively leverage historical exploration signals, and its optimization decision-making becomes much less reliable under limited evaluations. We further isolate the two LLM operators by retaining only the \textit{pattern explorer} (\textit{w/o Tuner}) and generating four new nodes at each step, or retaining only the \textit{knowledge tuner} (\textit{w/o Explorer}). In both cases, the DSE performance degrades even further. The \textit{tuner} ensures robust local refinements around promising designs, while the \textit{explorer} introduces structural diversity to prevent overly conservative search. Their collaboration is important for sustaining search quality across iterations.

\begin{figure}[t]
	\centering 
 \includegraphics[width=\linewidth]{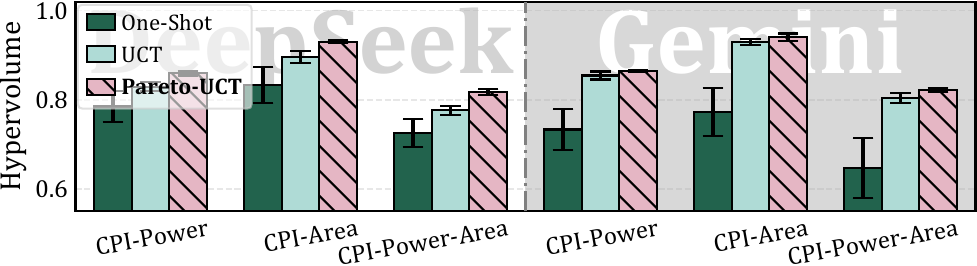} 
 \vspace{-20pt}
	\caption{Effectiveness of the proposed Pareto-UCT.}
	\label{fig:uct}
    \vspace{-10pt}
\end{figure}

\begin{table}[t]
\centering
\caption{Ablation study of MicroEvo under two evaluation budgets. Improvement (Imp.) is computed with respect to the full MicroEvo.}
\vspace{-12pt}
\label{tab:ablation_results}
\scriptsize
\setlength{\tabcolsep}{3.5pt}
\renewcommand{\arraystretch}{1.2}
\resizebox{\columnwidth}{!}{
\begin{tabular}{l|cccc|cccc}
\noalign{\hrule height 0.9pt}
\multirow{2}{*}{}
& \multicolumn{4}{c|}{Evaluation Budget = 20}
& \multicolumn{4}{c}{Evaluation Budget = 45} \\
\cline{2-9}
& HV $\uparrow$ & Imp. $\uparrow$ & ADRS $\downarrow$ & Imp. $\uparrow$
& HV $\uparrow$ & Imp. $\uparrow$ & ADRS $\downarrow$ & Imp. $\uparrow$ \\
\noalign{\hrule height 0.9pt}
\rowcolor[HTML]{E3F5EE}
\textbf{MicroEvo (Full)}
& \textbf{0.782} & 0.0\% & \textbf{0.077} & 0.0\%
& \textbf{0.817} & 0.0\% & \textbf{0.060} & 0.0\% \\
\hline
\textit{w/o AKA, SAD}
& 0.726 & -7.16\% & 0.111 & -44.73\%
& 0.762 & -6.73\% & 0.092 & -52.49\% \\

\textit{w/o Tuner}
& 0.717 & -8.33\% & 0.114 & -47.78\%
& 0.760 & -6.98\% & 0.087 & -44.60\% \\

\textit{w/o Explorer}
& 0.727 & -6.93\% & 0.101 & -31.71\%
& 0.746 & -8.63\% & 0.095 & -57.81\% \\
\hline
\textit{w/o AKA}
& 0.728 & -6.83\% & 0.110 & -42.95\%
& 0.771 & -5.54\% & 0.083 & -38.48\% \\

\textit{w/o Pareto}
& 0.746 & -4.58\% & 0.100 & -29.88\%
& 0.786 & -3.75\% & 0.074 & -23.99\% \\

\textit{w/o Pairwise}
& 0.735 & -6.01\% & 0.101 & -31.86\%
& 0.775 & -5.12\% & 0.081 & -35.39\% \\

\textit{w/o Utility}
& 0.737 & -5.69\% & 0.097 & -25.82\%
& 0.784 & -4.01\% & 0.073 & -22.05\% \\
\hline
\textit{w/o SAD}
& 0.727 & -7.03\% & 0.107 & -38.73\%
& 0.773 & -5.34\% & 0.082 & -35.86\% \\
\noalign{\hrule height 0.9pt}
\end{tabular}
}
\vspace{-5pt}
\end{table}

Within AKA, removing either Pareto analysis (\textit{w/o Pareto}) or pairwise analysis (\textit{w/o Pairwise}) yields worse results, highlighting the need for both a global optimization view and localized microarchitecture insights. The larger decrease observed in \textit{w/o Pairwise} suggests that extracting actionable optimization knowledge from concrete improvement cases is particularly critical. Moreover, replacing utility-aware memory selection with random retrieval (\textit{w/o Utility}) reduces performance. This result shows that effective exploration relies not only on accumulating experience, but also on retrieving the relevant insights at the right time. The utility mechanism enables flexible knowledge application, dynamically guiding microarchitecture optimization across changing search states. In the last row, we evaluate the effect of the state-aware directive. After removing the directive (\textit{w/o Directive}), the search quality drops again, verifying the importance of dynamic search state control.

\textbf{Effects of Pareto-UCT and iterative MCTS search.} Figure \ref{fig:uct} compares Pareto-UCT with a standard single-objective UCT policy. By introducing a multi-objective reward together with crowding-aware selection, Pareto-UCT more effectively identifies promising nodes within the MCTS tree. In contrast, single-objective UCT tends to over-exploit locally favorable regions and is more likely to get trapped in local optima. We also evaluate a one-shot LLM generation baseline that directly outputs 50 various design points. The MCTS-based MicroEvo achieves a clear advantage over the baseline, underscoring the importance of iterative learning in microarchitecture DSE, where effective decision-making should be continuously refined using feedback from prior evaluations. Moreover, the performance gap implies that existing LLMs cannot simply rely on memorizing optimal designs from their pre-training data.

\begin{figure}[t]
	\centering 
 \includegraphics[width=\linewidth]{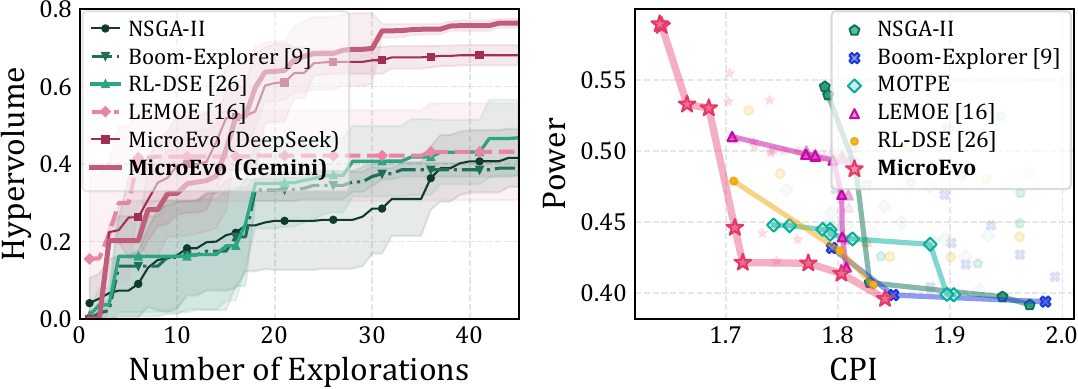} 
 \vspace{-20pt}
	\caption{Left: Hypervolume trajectory during the DSE process. Right: Explored Pareto fronts in the CPI-Power objective space.}
	\label{fig:xs_hv_space}
    \vspace{-9pt}
\end{figure}

\begin{figure}[t]
	\centering 
 \includegraphics[width=\linewidth]{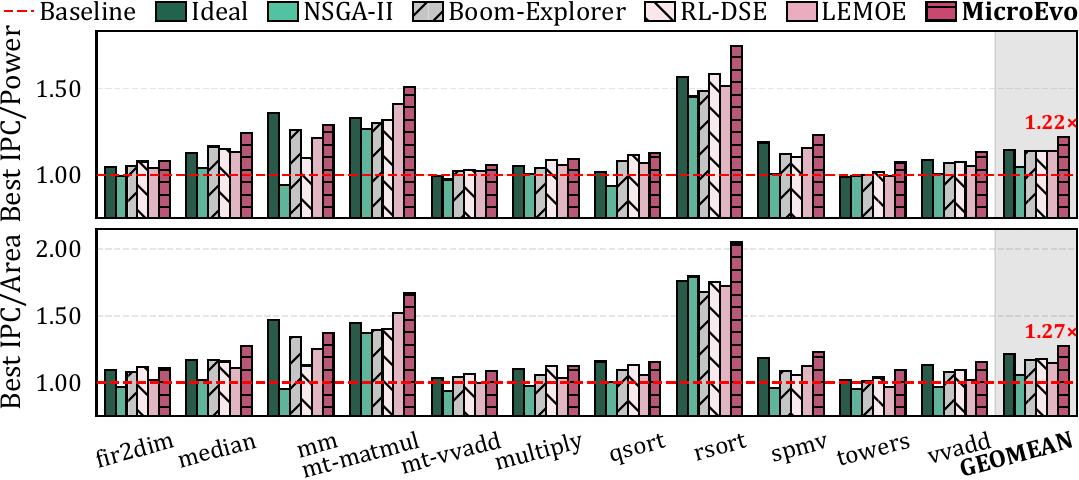} 
 \vspace{-20pt}
	\caption{Relative IPC/Power and IPC/Area improvement over the RTL-aligned XiangShan Kunminghu baseline across benchmarks.}
	\label{fig:benchmark}
\end{figure}

\subsection{Scaling to More Complex Microarchitecture}

In terms of portability, we further evaluate MicroEvo on the XiangShan Kunminghu processor using XS-GEM5. Compared with the Alpha21264-style OoO core, XiangShan features stronger parameter coupling and a more rugged search space. Figure \ref{fig:xs_hv_space} shows that MicroEvo still achieves the best HV trajectory and a broader, more concave Pareto front. Unguided sampling struggles to identify valuable points in vast combinatorial spaces, while surrogate-based methods are vulnerable to accumulated prediction errors. MicroEvo maintains strong sample efficiency through LLM-guided high-quality expansion and appropriate knowledge application, allowing it to scale to increasing microarchitectural complexity.

\textbf{Comparison with Expert-Crafted Design.} Figure~\ref{fig:benchmark} compares the best IPC/Power and IPC/Area explored, using the RTL-aligned Kunminghu configuration as baseline and the \textit{Ideal} hand-crafted XS-GEM5 design as an expert reference \cite{xsgem5}. MicroEvo consistently outperforms existing DSE methods, surpassing the manual design by 8.6\% in geometric mean energy efficiency. The advantage is most evident on \textit{mm}, \textit{rsort}, and \textit{spmv}, where efficient coordination across the memory hierarchy and control-flow resources matters more than simply increasing peak width. This improvement comes from the ability of MicroEvo to use optimization insights to guide the search and uncover effective yet non-obvious combinations of front-end, branch prediction, and memory-system resources.

\vspace{-5pt}

\section{Conclusion}
\label{sec:Conclusion}

In this paper, we have presented MicroEvo, a knowledge-guided LLM-MCTS framework for microarchitecture DSE. By integrating Pareto-aware tree search, experience accumulation, and state-aware expansion, MicroEvo outperforms prior methods in Pareto quality and sampling efficiency, and remains practical under a complex industrial-grade core. These results suggest that the value of LLMs in microarchitecture multi-objective optimization lies in turning evaluation feedback into more effective exploration decisions.

\section*{Acknowledgment}
This work is supported by the National Natural Science Foundation of China (Grant No.92464301, U25A6023), the National Key Research and Development Program (Grant No.2024YFB4405600), and the Key Research and Development Program of Jiangsu Province (Grant No.BG2024010).

\bibliographystyle{ACM-Reference-Format}
\bibliography{ref}

@INPROCEEDINGS{cpu_mobile,
  author={Halpern, Matthew and Zhu, Yuhao and Reddi, Vijay Janapa},
  booktitle={2016 IEEE International Symposium on High Performance Computer Architecture (HPCA)}, 
  title={Mobile CPU's rise to power: Quantifying the impact of generational mobile CPU design trends on performance, energy, and user satisfaction}, 
  year={2016},
  volume={},
  number={},
  pages={64-76}
  }

@ARTICLE{cpu_autodriving,
  author={Liu, Shaoshan and Liu, Liangkai and Tang, Jie and Yu, Bo and Wang, Yifan and Shi, Weisong},
  journal={Proceedings of the IEEE}, 
  title={Edge Computing for Autonomous Driving: Opportunities and Challenges}, 
  year={2019},
  volume={107},
  number={8},
  pages={1697-1716}
  }

@INPROCEEDINGS{boomexplorer,
  author={Bai, Chen and Sun, Qi and Zhai, Jianwang and Ma, Yuzhe and Yu, Bei and Wong, Martin D.F.},
  booktitle={2021 IEEE/ACM International Conference On Computer Aided Design (ICCAD)}, 
  title={BOOM-Explorer: RISC-V BOOM Microarchitecture Design Space Exploration Framework}, 
  year={2021},
  volume={},
  number={},
  pages={1-9}
  }

@inproceedings{rldse, 
title={Towards Automated RISC-V Microarchitecture Design with Reinforcement Learning}, 
volume={38}, 
number={1}, 
booktitle={2024 AAAI Conference on Artificial Intelligence (AAAI)}, 
author={Bai, Chen and Zhai, Jianwang and Ma, Yuzhe and Yu, Bei and Wong, Martin D. F.}, 
year={2024}, 
pages={12-20} 
}

@INPROCEEDINGS{LEMOE,
  author={Li, Jingyuan and Zhang, Jianrong and Li, Ye and Yin, Wenbo and Wang, Lingli},
  booktitle={2025 62nd ACM/IEEE Design Automation Conference (DAC)}, 
  title={LEMOE: LLM-Enhanced Multi-Objective Bayesian Optimization for Microarchitecture Exploration}, 
  year={2025},
  volume={},
  number={},
  pages={1-7},
  }

@INPROCEEDINGS{ChatA2,
  author={Zhu, Zhantong and Bai, Kangbo and Jia, Tianyu},
  booktitle={2026 31st Asia and South Pacific Design Automation Conference (ASP-DAC)}, 
  title={Chat-A2: An LLM-aided Design Space Exploration Framework for High-Performance CPU Design}, 
  year={2026},
  volume={},
  number={},
  pages={540-546},
  }

@ARTICLE{aldse,
  author={Zhai, Jianwang and Cai, Yici},
  journal={IEEE Transactions on Very Large Scale Integration (VLSI) Systems}, 
  title={Microarchitecture Design Space Exploration via Pareto-Driven Active Learning}, 
  year={2023},
  volume={31},
  number={11},
  pages={1727-1739},
  }

@ARTICLE{APPLE-DSE,
  author={Zhao, Xuyang and Gao, Tianning and Wu, Zheng and Bi, Zhaori and Yan, Changhao and Yang, Fan and Wang, Sheng-Guo and Zhou, Dian and Zeng, Xuan},
  journal={IEEE Transactions on Computer-Aided Design of Integrated Circuits and Systems}, 
  title={APPLE-DSE: Asynchronous Parallel Pareto Set Learning for Microarchitecture Design Space Exploration}, 
  year={2025},
  volume={44},
  number={7},
  pages={2765-2778},
  }

@inproceedings{ArchExplorer,
author = {Bai, Chen and Huang, Jiayi and Wei, Xuechao and Ma, Yuzhe and Li, Sicheng and Zheng, Hongzhong and Yu, Bei and Xie, Yuan},
title = {ArchExplorer: Microarchitecture Exploration Via Bottleneck Analysis},
year = {2023},
booktitle = {56th Annual IEEE/ACM International Symposium on Microarchitecture},
pages = {268–282},
numpages = {15},
series = {MICRO}
}

@article{ChatDSE,
author = {Tang, Mingxin and Chen, Wei and Wu, Lizhou and Huang, Libo and Zeng, Kun},
title = {ChatDSE: A Zero-Shot Microarchitecture Design Space Explorer Powered by GPT4.0},
year = {2025},
volume = {30},
number = {4},
journal = {ACM Trans. Des. Autom. Electron. Syst.},
numpages = {24},
}

@article{ChatArch,
author = {Wu, Zheng and Yang, Zhuochu and Yang, Zhuoyuan and Chen, Zihao and Shang, Li and Yang, Fan},
title = {ChatArch: A Knowledge-driven Graph-of-thought LLM Framework for Processor Architecture Optimization},
year = {2025},
volume = {31},
number = {2},
journal = {ACM Trans. Des. Autom. Electron. Syst.},
numpages = {26},
}

@INPROCEEDINGS{IT-DSE,
  author={Yu, Ziyang and Bai, Chen and Hu, Shoubo and Chen, Ran and He, Taohai and Yuan, Mingxuan and Yu, Bei and Wong, Martin},
  booktitle={2023 IEEE/ACM International Conference on Computer Aided Design (ICCAD)}, 
  title={IT-DSE: Invariance Risk Minimized Transfer Microarchitecture Design Space Exploration}, 
  year={2023},
  volume={},
  number={},
  pages={1-9},
  }

@inproceedings{MetaDSE,
author = {Xue, Runzhen and Wu, Hao and Yan, Mingyu and Xiao, Ziheng and Ye, Xiaochun and Fan, Dongrui},
title = {MetaDSE: A Few-Shot Meta-Learning Framework for Cross-Workload CPU Design Space Exploration},
year = {2025},
booktitle = {2025 62nd Annual ACM/IEEE Design Automation Conference},
numpages = {7},
series = {DAC}
}

@INPROCEEDINGS{trendse,
  author={Wang, Duo and Yan, Mingyu and Teng, Yihan and Han, Dengke and Dang, Haoran and Ye, Xiaochun and Fan, Dongrui},
  booktitle={2023 IEEE/ACM International Conference on Computer Aided Design (ICCAD)}, 
  title={A Transfer Learning Framework for High-Accurate Cross-Workload Design Space Exploration of CPU}, 
  year={2023},
  volume={},
  number={},
  pages={1-9},
  }

@INPROCEEDINGS{modse,
  author={Wang, Duo and Yan, Mingyu and Liu, Xin and Zou, Mo and Liu, Tianyu and Li, Wenming and Ye, Xiaochun and Fan, Dongrui},
  booktitle={2023 60th ACM/IEEE Design Automation Conference (DAC)}, 
  title={A High-accurate Multi-objective Exploration Framework for Design Space of CPU}, 
  year={2023},
  volume={},
  number={},
  pages={1-6},
  }

@inproceedings{motpe,
author = {Ozaki, Yoshihiko and Tanigaki, Yuki and Watanabe, Shuhei and Onishi, Masaki},
title = {Multiobjective tree-structured parzen estimator for computationally expensive optimization problems},
year = {2020},
booktitle = {2020 Genetic and Evolutionary Computation Conference},
pages = {533–541},
numpages = {9},
series = {GECCO}
}

@inproceedings{EoH,
author = {Liu, Fei and Tong, Xialiang and Yuan, Mingxuan and Lin, Xi and Luo, Fu and Wang, Zhenkun and Lu, Zhichao and Zhang, Qingfu},
title = {Evolution of heuristics: towards efficient automatic algorithm design using large language model},
year = {2024},
booktitle = {2024 41st International Conference on Machine Learning (ICML)},
numpages = {23},
}

@article{FunSearch,
  title={Mathematical discoveries from program search with large language models},
  author={Romera-Paredes, Bernardino and Barekatain, Mohammadamin and Novikov, Alexander and Balog, Matej and Kumar, M Pawan and Dupont, Emilien and Ruiz, Francisco JR and Ellenberg, Jordan S and Wang, Pengming and Fawzi, Omar and others},
  journal={Nature},
  volume={625},
  number={7995},
  pages={468--475},
  year={2024},
}

@inproceedings{ReEvo,
 author = {Ye, Haoran and Wang, Jiarui and Cao, Zhiguang and Berto, Federico and Hua, Chuanbo and Kim, Haeyeon and Park, Jinkyoo and Song, Guojie},
 booktitle = {Advances in Neural Information Processing Systems},
 pages = {43571--43608},
 title = {ReEvo: Large Language Models as Hyper-Heuristics with Reflective Evolution},
 volume = {37},
 year = {2024}
}

@article{CogMCTS,
  title={CogMCTS: A Novel Cognitive-Guided Monte Carlo Tree Search Framework for Iterative Heuristic Evolution with Large Language Models}, 
  author={Hui Wang and Yang Liu and Xiaoyu Zhang and Chaoxu Mu},
  year={2025},
  journal={arXiv preprint arXiv:2512.08609}
}

@inproceedings{HiFo-Prompt,
  title={Hifo-prompt: Prompting with hindsight and foresight for llm-based automatic heuristic design},
  author={Zhong, Mengyuan and Shi, Jialong and Sun, Jianyong and Fan, Ye},
  booktitle={International Conference on Learning Representations (ICLR)},
  volume={2026},
  pages={117102--117143},
  year={2026}
}

@ARTICLE{EC_Review,
  author={Wu, Xingyu and Wu, Sheng-Hao and Wu, Jibin and Feng, Liang and Tan, Kay Chen},
  journal={IEEE Transactions on Evolutionary Computation}, 
  title={Evolutionary Computation in the Era of Large Language Model: Survey and Roadmap}, 
  year={2025},
  volume={29},
  number={2},
  pages={534-554},
}

@inproceedings{MCTS-AHD,
  title={Monte Carlo Tree Search for Comprehensive Exploration in LLM-Based Automatic Heuristic Design},
  author={Zheng, Zhi and Xie, Zhuoliang and Wang, Zhenkun and Hooi, Bryan},
  booktitle={2025 International Conference on Machine Learning (ICML)},
  pages={78338--78373},
  year={2025}
}

@inproceedings{UCT,
  title={Bandit based monte-carlo planning},
  author={Kocsis, Levente and Szepesv{\'a}ri, Csaba},
  booktitle={European conference on machine learning},
  pages={282--293},
  year={2006},
  organization={Springer}
}

@article{mcts_review,
  title={Monte Carlo tree search: A review of recent modifications and applications},
  author={{\'S}wiechowski, Maciej and Godlewski, Konrad and Sawicki, Bartosz and Ma{\'n}dziuk, Jacek},
  journal={Artificial Intelligence Review},
  volume={56},
  number={3},
  pages={2497--2562},
  year={2023},
  publisher={Springer}
}

@inproceedings{analysis_isca,
author = {Karkhanis, Tejas S. and Smith, James E.},
title = {Automated design of application specific superscalar processors: an analytical approach},
year = {2007},
booktitle = {2007 34th Annual International Symposium on Computer Architecture},
pages = {402–411},
numpages = {10},
series = {ISCA}
}

@INPROCEEDINGS{analysis_micro,
  author={Sun, Guangyu and Hughes, Christopher and Kim, Changkyu and Zhao, Jishen and Xu, Cong and Xie, Yuan and Chen, Yen-Kuang},
  booktitle={2011 38th Annual International Symposium on Computer Architecture (ISCA)}, 
  title={Moguls: A model to explore the memory hierarchy for bandwidth improvements}, 
  year={2011},
  volume={},
  number={},
  pages={377-388},
  }

@ARTICLE{xiangshan,
  author={Wang, Kaifan and Chen, Jian and Xu, Yinan and Yu, Zihao and He, Wei and Tang, Dan and Sun, Ninghui and Bao, Yungang},
  journal={IEEE Micro}, 
  title={XiangShan: An Open Source Project for High-Performance RISC-V Processors Meeting Industrial-Grade Standards}, 
  year={2025},
  volume={45},
  number={3},
  pages={49-57}
  }

@INPROCEEDINGS{samsung,
  author={Grayson, Brian and Rupley, Jeff and Zuraski, Gerald Zuraski and Quinnell, Eric and Jiménez, Daniel A. and Nakra, Tarun and Kitchin, Paul and Hensley, Ryan and Brekelbaum, Edward and Sinha, Vikas and Ghiya, Ankit},
  booktitle={2020 ACM/IEEE 47th Annual International Symposium on Computer Architecture (ISCA)}, 
  title={Evolution of the Samsung Exynos CPU Microarchitecture}, 
  year={2020},
  volume={},
  number={},
  pages={40-51},
  }

@inproceedings{chipmind, 
title={ChipMind: Retrieval-Augmented Reasoning for Long-Context Circuit Design Specifications}, 
volume={40}, 
number={2}, 
booktitle={2026 AAAI Conference on Artificial Intelligence (AAAI)}, 
author={Xing, Changwen and Wong, SamZaak and Wan, Xinlai and Lu, Yanfeng and Zhang, Mengli and Ma, Zebin and Qi, Lei and Li, Zhengxiong and Guan, Nan and Jiang, Zhe and Wang, Xi and Yang, Jun}, 
year={2026}, 
pages={1337-1345} 
}

@inproceedings{ChatCPU,
author = {Wang, Xi and Wan, Gwok-Waa and Wong, Sam-Zaak and Zhang, Layton and Liu, Tianyang and Tian, Qi and Ye, Jianmin},
title = {ChatCPU: An Agile CPU Design and Verification Platform with LLM},
year = {2024},
booktitle = {2024 61st ACM/IEEE Design Automation Conference (DAC)},
numpages = {6},
}

@INPROCEEDINGS{Wavefront-MCTS,
  author={Hu, Yong and Mueller-Gritschneder, Daniel and Schlichtmann, Ulf},
  booktitle={2018 IEEE/ACM International Conference on Computer-Aided Design (ICCAD)}, 
  title={Wavefront-MCTS: Multi-objective Design Space Exploration of NoC Architectures based on Monte Carlo Tree Search}, 
  year={2018},
  volume={},
  number={},
  pages={1-8},
  }

@INPROCEEDINGS{profiling_gem5,
  author={Umeike, Johnson and Patel, Neel and Manley, Alex and Mamandipoor, Amin and Yun, Heechul and Alian, Mohammad},
  booktitle={2023 IEEE International Symposium on Performance Analysis of Systems and Software (ISPASS)}, 
  title={Profiling GEM5 Simulator}, 
  year={2023},
  volume={},
  number={},
  pages={103-113},
  }

@inproceedings{mcpat,
author = {Li, Sheng and Ahn, Jung Ho and Strong, Richard D. and Brockman, Jay B. and Tullsen, Dean M. and Jouppi, Norman P.},
title = {McPAT: an integrated power, area, and timing modeling framework for multicore and manycore architectures},
year = {2009},
booktitle = {42nd Annual IEEE/ACM International Symposium on Microarchitecture},
pages = {469–480},
numpages = {12},
series = {MICRO}
}

@article{gem5,
author = {Binkert, Nathan and Beckmann, Bradford and Black, Gabriel and Reinhardt, Steven K. and Saidi, Ali and Basu, Arkaprava and Hestness, Joel and Hower, Derek R. and Krishna, Tushar and Sardashti, Somayeh and Sen, Rathijit and Sewell, Korey and Shoaib, Muhammad and Vaish, Nilay and Hill, Mark D. and Wood, David A.},
title = {The GEM5 simulator},
year = {2011},
volume = {39},
number = {2},
journal = {SIGARCH Comput. Archit. News},
pages = {1–7},
numpages = {7}
}

@misc{riscv-test,
  title={Official RISC-V Benchmark Suites},
  year = {2026},
  url = {https://github.com/riscv-software-src/riscv-tests},
}

@article{hv,
author = {Guerreiro, Andreia P. and Fonseca, Carlos M. and Paquete, Lu\'{\i}s},
title = {The Hypervolume Indicator: Computational Problems and Algorithms},
year = {2021},
volume = {54},
number = {6},
journal = {ACM Comput. Surv.},
numpages = {42},
}

@misc{xsgem5,
  title={XS-GEM5},
  year = {2026},
  url = {https://github.com/OpenXiangShan/GEM5},
}

@article{chatsva,
  title={ChatSVA: Bridging SVA Generation for Hardware Verification via Task-Specific LLMs},
  author={Fu, Lik Tung and Zhou, Jie and Ren, Shaokai and Zhang, Mengli and Xiong, Jia and Jiang, Hugo and Guan, Nan and Wang, Xi and Yang, Jun},
  journal={arXiv preprint arXiv:2604.02811},
  year={2026}
}

@INPROCEEDINGS{amd_processor,
  author={Naffziger, Samuel and Beck, Noah and Burd, Thomas and Lepak, Kevin and Loh, Gabriel H. and Subramony, Mahesh and White, Sean},
  booktitle={2021 ACM/IEEE 48th Annual International Symposium on Computer Architecture (ISCA)}, 
  title={Pioneering Chiplet Technology and Design for the AMD EPYC™ and Ryzen™ Processor Families : Industrial Product}, 
  year={2021},
  volume={},
  number={},
  pages={57-70}
}

@inproceedings{chathls,
    title = "{C}hat{HLS}: Towards Systematic Design Automation and Optimization for High-Level Synthesis",
    author = "Li, Runkai  and
      Xiong, Jia  and
      He, Xiuyuan  and
      Zhao, Jieru  and
      Lv, Jiaqi  and
      Fang, Haowen  and
      Qi, Lei  and
      Wang, Xi",
    booktitle = "2026 64th Annual Meeting of the {A}ssociation for {C}omputational {L}inguistics (Volume 1: Long Papers) (ACL)",
    year = "2026",
    pages = "20996--21015"
}

@inproceedings{Fixme,
  title = {{FIXME}: Towards End-to-End Benchmarking of {LLM}-Aided Design Verification},
  author={Wan, Gwok-Waa and Wong, SamZaak and Su, Shengchu and Niu, Chenxu and Wang, Ning and Wan, Xinlai and Chen, Qixiang and Xing, Mengnv and Zhang, Jingyi and Ye, Jianmin and others},
  booktitle={2026 AAAI Conference on Artificial Intelligence (AAAI)},
  volume={40},
  number={2},
  pages={1087--1095},
  year={2026}
}

@article{idse,
  title = {iDSE: Navigating Design Space Exploration in High-Level Synthesis Using {LLMs}},
  author={Li, Runkai and Xiong, Jia and Wang, Xi},
  journal={arXiv preprint arXiv:2505.22086},
  year={2025}
}

@article{chatmodel,
  title = {{ChatModel}: Automating Reference Model Design and Verification with {LLMs}},
  author={Ye, Jianmin and Liu, Tianyang and Tian, Qi and Su, Shengchu and Jiang, Zhe and Wang, Xi},
  journal={arXiv preprint arXiv:2506.15066},
  year={2025}
}

\end{document}